\documentclass{article} 
\usepackage{iclr2024_conference,times}

\usepackage{amsmath,amsfonts,bm}

\def\eqref#1{equation~\ref{#1}}

\def\1{\bm{1}}

\DeclareMathAlphabet{\mathsfit}{\encodingdefault}{\sfdefault}{m}{sl}
\SetMathAlphabet{\mathsfit}{bold}{\encodingdefault}{\sfdefault}{bx}{n}

\usepackage{multirow}
\usepackage{float}

\usepackage[utf8]{inputenc} 
\usepackage[T1]{fontenc}    
\usepackage{url}            
\usepackage{booktabs}       
\usepackage{amsfonts}       
\usepackage{nicefrac}       
\usepackage{microtype}      

\usepackage{mathtools}
\usepackage{natbib}
\usepackage{xcolor}
\usepackage{url}
\usepackage{amsmath}
\usepackage{amssymb}
\usepackage{amsthm}
\usepackage{graphicx}
\usepackage{subcaption}
\usepackage{nicefrac}
\usepackage{appendix}
\usepackage{enumitem}
\usepackage{arydshln}
\definecolor{darker}{rgb}{0,0.15,0.7}
\usepackage[colorlinks, urlcolor=darker, citecolor=darker, linkcolor=darker]{hyperref} 
\setcitestyle{round}
\newcommand{\llama}{Llama2-70B}
\newcommand{\ourllm}{GPT-43B}

\usepackage{tikz}

\title{Retrieval meets Long Context Large Language Models}

\author{Peng Xu$^{\dagger}$, Wei Ping$^{\dagger}$,  Xianchao Wu, Lawrence McAfee \vspace{.8mm} \\ 
{\bf  Chen Zhu,  Zihan Liu,  Sandeep Subramanian, Evelina Bakhturina} \vspace{.8mm} \\ 
{\bf Mohammad Shoeybi, Bryan Catanzaro} \vspace{1.2mm} \\
NVIDIA \vspace{.5mm} \\
$^{\dagger}$\texttt{\{pengx, wping\}@nvidia.com} 
}

\iclrfinalcopy 

\begin{document}

\maketitle

\begin{abstract}
Extending the context window of large language models~(LLMs) is getting popular recently, while the solution of augmenting LLMs with retrieval has existed for years.
The natural questions are: \emph{i) Retrieval-augmentation versus long context window, which one is better for downstream tasks?}~\emph{ii) Can both methods be combined to get the best of both worlds?}
In this work, we answer these questions by studying both solutions using two state-of-the-art pretrained LLMs, i.e., a proprietary 43B GPT and \llama. 
Perhaps surprisingly, we find that LLM with 4K context window using simple retrieval-augmentation at generation can achieve comparable performance to finetuned LLM with 16K context window via \emph{positional interpolation} on long context tasks, while taking much less computation.
More importantly, we demonstrate that retrieval can significantly improve the performance of LLMs regardless of their extended context window sizes. 
Our best model, retrieval-augmented {\llama} with 32K context window, outperforms GPT-3.5-turbo-16k and Davinci003 in terms of average score on nine long context tasks including question answering, query-based summarization, and in-context few-shot learning tasks.
It also outperforms its non-retrieval {\llama}-32k baseline by a margin, while being much faster at generation.
Our study provides general insights on the choice of retrieval-augmentation versus long context extension of LLM for practitioners.
\end{abstract}

\section{Introduction}
\label{sec:intro}
The long context large language models~(LLM) have recently received a lot of attention in production~\citep[e.g.,][]{claud_100k,openai_32k},  research community~\citep[e.g.,][]{chen2023extending, liu2023lost, tworkowski2023focused}, and open source community~\citep[e.g.,][]{kaiokendev2023}.
Although the \emph{approximate} attention methods have been studied for years~\citep[e.g.,][]{Tay2022efficient} (due to the quadratic time and memory complexities of self-attention mechanism in sequence length), 
the recent advance for long context LLMs with \emph{exact} attention is mainly driven by the development of faster GPU with more memory and memory-efficient exact attention~\citep{dao2022flashattention, dao2023flashattention}. 

An alternative and long-standing solution for handling long context is \emph{retrieval}.
Specifically, the LLMs only read relevant context retrieved from a  standalone retriever~\citep[e.g.,][]{karpukhin2020dense, wang2022text, lin2023train}, which is much easier to scale~\footnote{The dense embedding retriever can easily retrieve context from billions of tokens using the fast similarity search library~\citep{faiss}.} and runs orders of magnitudes faster than LLMs for selecting relevant context.   
Conceptually, the retrieval-augmented decoder-only LLM can be viewed as applying the sparse attention over its long context window, where the sparsity pattern is not predefined as \citet{child2019generating} but determined by the standalone retriever. In other words, unretrieved context is treated as irrelevant and has zero-valued attention weights.

Given the surge of interest in long context LLM research and much more required computation at inference~\footnote{For example, the price of GPT-4 with 32k context length is twice the 8k context model.}, it is still unclear for practitioners whether extending the context window of LLM provides higher accuracy than the retrieval-augmentation for downstream tasks with informative queries.
Moreover, it would be compelling if we could combine the strength of both methods and achieve even higher accuracies. 
In this work, we attempt to answer the above questions through a comprehensive study.

Specifically, we make the following contributions: 
\vspace{-0.4em}
\begin{enumerate}[leftmargin=3.1em]
    \item We perform comprehensive study using two state-of-the-art LLMs, a proprietary 43B pretrained GPT and 
{\llama} ~\citep{touvron2023llama2} on 9 downstream long context tasks, including single and multi document question answering~(QA), query-based summarization, and in context few-shot learning tasks.
    \item We demonstrate that retrieval-augmentation significantly improves the performance of 4K context LLMs. Perhaps surprisingly, we find this simple retrieval-augmented baseline can perform comparable to 16K long context LLMs, i.e., average score 29.32 vs. 29.45 by using {\ourllm}, and 36.02 vs. 36.78 by using {\llama}, while using much less computation.
    \item Furthermore, we demonstrate that the performance of long context LLM~(i.e., 16K or 32K) can still be improved by retrieval, especially for the larger {\llama}. 
    As a result, our best model, retrieval augmented {\llama}-32k-ret with 32K context window~(avg. score 43.6), outperforms GPT-3.5-turbo-16k~(avg. score 42.8) and Davinci-003 in terms of average score. 
    It also largely outperforms its non-retrieval {\llama}-32k baseline~(avg. score 40.9), while can be much faster at generation~(e.g., 4$\times$ faster on NarrativeQA).
\vspace{-0.4em}
\end{enumerate}

We organize the rest of the paper as follows.
We discuss related work in Section~\ref{sec:related_work}, and present the experimental setup in Section~\ref{sec:setup}.
We report results in Section~\ref{sec:results} and conclude the paper in Section~\ref{sec:conclusion}.

\section{Related Work}
\label{sec:related_work}
In this section, we discuss the related work in long context LLM, efficient attention methods, and retrieval-augmented language models.

\subsection{Parallel work}
When we are preparing this manuscript, we notice that a concurrent work~\citep{bai2023longbench}~(arXived on 28 Aug 2023) also studies the impact of retrieval on long context LLM, including black-box model GPT-3.5-Turbo-16k~\citep{chatgpt}, white-box model Llama2-7B-chat-4k~\citep{touvron2023llama2}, and ChatGLM2-6B-32k~\citep{zeng2022glm}.
Different from our findings, they find that retrieval is only helpful for Llama2-7B-chat-4k with 4K context window, but not helpful for long context model, i.e., GPT-3.5-Turbo-16k and ChatGLM2-6B-32k.
We hypothesize the major reasons are: \emph{i)} it is challenging to do controlled experiments using black-box APIs,
\emph{ii)} the white-box LLMs used in their study are relatively small, thus they have limited zero-shot capability of incorporating context through retrieval.
Our conclusions are drawn from much larger LLMs. In particular, our best long context model {\llama}-32k performs as well as Davinci003 and GPT-3.5-turbo-16k, while it can still be further enhanced by retrieval~(see Table~\ref{tab:compare-to-chatgpt}).

\subsection{Long Context Large Language Models}
Over the past few years, pretraining large language models~(LLMs) with long context window becomes a viable solution thanks to faster GPU with more memory and memory-efficient exact attention~\citep[e.g.,][]{dao2022flashattention}. 
For example, the context window for pretrained LLM have been increased from 1024 of GPT-2~\citep{radford2019language}, 2048 of GPT-3~\citep{brown2020language}, 4096 of Llama~2~\citep{touvron2023llama2}, to 8192 of GPT-4~\citep{gpt4}.
However, further extending the context window in pretraining can be challenging, because, \emph{i)} pretraining LLM from scratch with long context~(e.g., >16K tokens) is very expensive due to the quadratic time and memory complexities of exact attention, and \emph{ii)} most of documents in pretraining corpus~(e.g., Common Crawl) are relatively short.

Most recently, researchers start to extend the context window of LLMs with continued training or fine-tuning~\citep[e.g.,][]{kaiokendev2023, Nijkamp2023xgen, chen2023extending, tworkowski2023focused, mohtashami2023landmark}.
\citet{tworkowski2023focused} introduced LongLLaMA by fine-tuning the 3B and 7B OpenLLaMA checkpoints with contrastive training on 8K context length.
Landmark attention~\citep{mohtashami2023landmark} extends the context length of LLaMA 7B from 4K to 32K by introducing ``landmark tokens'' to represent  blocks of the context and fine-tuning the attention to use landmark tokens for selecting relevant blocks.
\citet{chen2023extending} and \citet{kaiokendev2023} introduced \emph{positional interpolation} to extend the context window sizes of RoPE-based~\citep{su2021roformer} pretrained LLMs.  
In particular, \citet{chen2023extending} demonstrates promising results on LLaMA 7B to 65B~\citep{touvron2023llama} with minimal fine-tuning effort~(within 1000 steps).
ALiBi~\citep{press2021train} extrapolates context window length by removing the positional embeddings while simply biasing the key-query attention scores with a linear penalty that is proportional to their distance, so one does not need finetuning for context window extrapolation.
\citet{ratner2023parallel} chunks long context into multiple sub-windows and re-use the positional embeddings across these windows, thus can handle longer context  without any further finetuning.
In this work, we apply \emph{positional interpolation} method to extend the 4K context window of a proprietary 43B pretrained LLM and  {\llama} ~\citep{touvron2023llama2} to 16K and 32K, as they both use rotary position embedding at pretraining.
In terms of evaluation, we focus on downstream task performance~\citep[e.g.,][]{shaham2023zeroscrolls, bai2023longbench} after instruction tuning~\citep{wei2021finetuned}. 

There are other studies showing the interplay between retrieval-augmentation and long context LLM. \citet{liu2023lost} performs the black-box evaluation for the long context capability of existing LLM products, including ChatGPT~3.5~\citep{chatgpt}, GPT-4~\citep{gpt4}, Claude~\citep{claud_100k}, in retrieval-augmented setting, and identify the ``lost in the middle'' phenomenon in these models.

\subsection{Efficient Attention Methods}
In previous study, many approximate attention methods~\citep{Tay2022efficient} have been introduced for dealing with the quadratic complexity of self-attention that becomes a computational bottleneck for long context.
They can be  grouped into the following categories: \emph{i)} Sparse attention mechanisms with predefined sparsity patterns~\citep[e.g.,][]{child2019generating, parmar2018image, ho2019axial, beltagy2020longformer, zaheer2020big, zhu2021long},
\emph{ii)}~recurrence-based method~\citep{dai2019transformer, bulatov2022recurrent},
\emph{iii)} low-rank projection attention~\citep[e.g.,][]{wang2020linformer, xiong2021nformer, tay2021synthesizer, zhu2021long}, 
\emph{iv)} memory-based mechanisms~\citep[e.g.,][]{rae2019compressive, liu2018generating},
\emph{v)} similarity and clustering based methods~\citep[e.g.,][]{kitaev2020reformer, tay2020sparse, roy2021efficient}.
These approximate methods introduce inductive bias~(e.g., predefined sparsity) that can fit well for specific domain, but may reduce model quality in general LLM training.

Most recently, FlashAttention~\citep{dao2022flashattention, dao2023flashattention} is introduced to speed up the exact attention computation by accounting for reads and writes between levels of GPU memory. FlashAttention is particularly useful for handling longer sequences.

\subsection{Retrieval-augmented Language Models}
Retrieval has been integrated into language models for years to improve perplexity~\citep{borgeaud2022improving, wang2023shall}, factual accuracy~\citep{nakano2021webgpt}, downstream task accuracy~\citep{guu2020retrieval, izacard2021leveraging, izacard2022few, lewis2020retrieval}, and in-context learning capability~\citep{huang2023raven}.
Combined with a standalone retriever~\citep{karpukhin2020dense, wang2022text, lin2023train}, retrieval-augmented LLM is well established for handling question answering with long document and in open-domain.
In previous study, language models have been augmented with retrieval at inference~\citep{khandelwal2019generalization, yogatama2021adaptive}, fine-tuning~\citep{izacard2022few, lewis2020retrieval, guu2020retrieval}, and pretraining~\citep{borgeaud2022improving, izacard2022few, wang2023shall}. 
There are also methods that try to integrate LLM and retriever in a single model and build the end-to-end solution 
\citep[e.g., ][]{jiang2022retrieval, shi2023replug}.
However, most of previous works mainly study retrieval-augmentation for LLMs that have around 10 billion parameters, except a few recent ones~\citep[e.g.,][]{shi2023replug}. 

In this work, we focus on decoder-only LLMs with 43B and 70B parameters trained on trillions of tokens, because the LLMs at such scale exhibit strong zero-shot capability to incorporate context after instruction tuning~\citep{wei2021finetuned, wei2022emergent}.

\section{Experimental Setup}
\label{sec:setup}
In this section, we present the details of our experimental setup.
\subsection{Large Language Models}
We focus on comparing the zero-shot capability of integrating long context information for generative QA or summarization tasks via retrieval or LLM's own self-attention mechanism.
In contrast to most existing works that focus on relatively small models~(e.g., 3B or 7B)~\citep{kaiokendev2023, Nijkamp2023xgen, tworkowski2023focused, mohtashami2023landmark}, we gather the insights by exploring model sizes that are larger than 40B after instruction tuning, as previous study suggests that instruction tuning becomes effective when the decoder-only LLM has around 50B parameters~\citep{wei2021finetuned, wei2022emergent}.

Specifically, we experimented with two pretrained GPT models, a proprietary Nemo {\ourllm} and {\llama}. 
{\ourllm} is a 43 billion parameter model that is trained with 1.1T tokens with 70\% English corpus and the other 30\% for multilingual and code data. For the English pretraining corpus, {\ourllm} used  Common Crawl
web archive (WARC), Wikipedia, Reddit, Books, Gutenberg, ArXiv, StackExchange, PubMed, etc. It contains 48 layers with the hidden dimension of 8,192. It is trained with a sequence length of 4,096 and RoPE embeddings \citep{su2021roformer}. The other {\llama} is a public available 70B GPT model trained on 2T tokens using around 90\% English data. It contains 80 layers with the hidden dimension of 8,192. It also has the context window size of 4,096 and trained with RoPE embeddings.

\subsection{Datasets and Metrics}
In this study, we include seven datasets ranging from single document QA, multi document QA, to query-based summarization for our zero shot evaluations. 
Specifically, we include four datasets from the validation set of the Scroll benchmark~\citep{shaham-etal-2022-scrolls}.
\begin{itemize}[leftmargin=2.1em]
    \item \textbf{QMSum (QM)} ~\citep{zhong-etal-2021-qmsum} is a query-based summarization dataset, consisting of meetings' transcripts and their corresponding summaries from multiple domains such as academic, industrial product. In this task, a meeting dialogue transcript is given, and a question to summarize certain topic is raised about the dialogue, such as ``what is agreed between them''. The answer generally contains a few sentences. 
    \item  \textbf{Qasper (QASP)}~\citep{dasigi-etal-2021-dataset-qasper} is a question answering dataset over NLP papers filtered from the Semantic Scholar Open Research Corpus (S2ORC) ~\citep{lo-etal-2020-s2orc}. Qasper contains abstractive, extractive, and yes/no questions, as well as unanswerable ones. In this task, one script is provided together with an information seeking question, such as ``which multilingual approaches do they compare with?''. A model needs to give a short answer by reasoning over the given context.
    \item \textbf{NarrativeQA (NQA)} ~\citep{10.1162/tacl_a_00023-narrativeqa} is an established question answering dataset over entire books from Project Gutenberg\footnote{\url{https://www.gutenberg.org/}} and movie scripts from a list of websites.   In this task, the given passage is transcribed from books and is usually noisy. A model is required to generate a short phrase by reasoning over the long and noisy text. 
    \item \textbf{QuALITY (QLTY)}~\citep{pang-etal-2022-quality} is a question answering dataset over stories and articles collected from several resources, such as Project Gutenberg and the Open American National Corpus\footnote{\url{https://anc.org/}}. Different from all the other tasks, this is a multi-choices dataset and a model is required to select one among four given choices.  
\end{itemize}
We take another three datasets from LongBench~\citep{bai2023longbench}.
\begin{itemize}[leftmargin=2.1em]
    \item \textbf{HotpotQA (HQA)}~\citep{yang-etal-2018-hotpotqa} is a Wikipedia-based question-answer dataset. Different from above single hot datasets, HQA is a multi-hop dataset where multiple supporting documents are required to be read for answering and reasoning and the questions are diverse and not constrained to any pre-existing knowledge bases. 
    \item \textbf{MuSiQue (MSQ)}~\citep{trivedi-etal-2022-musique} is another multi-hop question answering dataset.  Compared to HQA, MSQ requires connected reasoning by reducing potential reasoning shortcuts, minimizing train-test leakage, and including harder distractor contexts. Thus, MSQ is much harder task than HQA and significantly less cheatable. 
    \item \textbf{MultiFieldQA-en (MFQA)}~\citep{bai2023longbench} was manually curated to better test the model’s long context understanding ability across diverse fields. The evidences from multiple sources, including legal documents, government reports, encyclopedias, and academic papers, are fairly randomly located in the documents to avoid biases that might occur at the beginning or ending of the documents. 
\end{itemize}
The full details of the dataset can be found in Table \ref{tab:7datasets-statistics}. We can see that our evaluation datasets have a wide range of average document length from 4.9k (QASP) to 84k (NQA). Therefore, for the baseline model without retrieval, we truncate the document accordingly to fit into the input sequence length.

\begin{table}[t]\small
\centering
\begin{tabular}{lrrrrrrr}
\toprule
  & QM  & QASP & NQA & QLTY  &MSQ &HQA &MFQA \\
  \midrule
\# of samples & 200 & 1,726  & 2,000  & 2,000  & 200 &  200 & 150 \\
avg doc length &  14,140 & 4,912 & 84,770 & 6,592 & 16,198 & 13,319 & 7,185 \\
avg top-5 chunks & 2,066 & 2,071 & 2,549 & 2,172 & 2,352 & 2,322 & 2,385 \\
avg top-10 chunks & 4,137 & 3,716 & 5,125 & 4,018 & 4,644 & 4,554 & 4,305 \\
avg top-20 chunks & 8,160 & 4,658 & 10,251 & 5,890 & 9,133 & 8,635 & 6,570 \\
\bottomrule
\end{tabular}
\caption{Statistics of seven datasets used for zero-shot evaluation. All lengths are counted by the number of tokens using {\llama} tokenizer, and ``avg top N chunks" denotes the average number of tokens from the top N retrieved chunks. Figure~\ref{fig:7datasets-top5-ctxs-boxplot} gives more details.}
\label{tab:7datasets-statistics}
\end{table}

Following the official metrics, we report the geometric mean of ROUGE scores (i.e., ROUGE-1/2/L)~\citep{lin-2004-rouge} for QM, the exact matching (EM) score for QLTY, and F1 scores for the remaining five datasets QASP, NQA, MSQ, HQA and MFQA.

\subsection{Context Window Extension}
We extend the context window length with position interpolation method~\citep{chen2023extending}, as it is simple and effective for RoPE embeddings. 
We extend the 4K context window to 16K for 
{\ourllm}. For {Llama2}, we extend its 4K context window to 32k for {Llama2-7B} and both 16K and 32K for {\llama}. 
We follow \citet{chen2023extending} and finetune both LLMs on the Pile dataset~\citep{DBLP:journals/corr/abs-2101-00027-pile-dataset}
with batch size as 128, constant learning rate of 5e-6 to adapt the position embeddings.

\subsection{Retrieval}
For the retriever, we experimented with three retrievers: 1) \emph{Dragon} ~\citep{lin2023train} as it achieves state-of-the-art results on both supervised and zero-shot information retrieval benchmarks~\citep{thakur2021beir}. Dragon is a dual encoder model that consists of a query encoder and a context encoder. 2) a widely used \emph{Contriever} model ~\citep{izacard2021contriever}.
Following the MoCo technique~\citep{He_2020_CVPR_MoCo}, Contriever used a simple contrastive learning framework to pre-train models for information retrieval. It was trained without supervision and achieved competitive results with BM25 for R@100 on the BEIR benchmark~\citep{thakur2021beir}, and 3) \emph{OpenAI embedding}\footnote{\url{https://platform.openai.com/docs/guides/embeddings}}. For the OpenAI embedding model, we use the latest ``text-embedding-ada-002'' as recommended by OpenAI. It accepts 8,191 maximum input tokens for one sequence with an output vector of 1,536 dimensions. The cosine similarities are then computed between the questions and the list of contexts for retrieval ranking. 

To use these retrievers, we first chunk each context document with 300 words, and then we encode both the questions and all chunks independently with corresponding encoders. The most relevant N chunks, ranked by the dot product of the question embedding and chunk embedding, are then concatenated together (following the left to right order from the most relevant to least relevant) as the context of the prompt for generation. Table \ref{tab:7datasets-statistics} shows the statistics of the top N retrieved chunks while Figure \ref{fig:7datasets-top5-ctxs-boxplot} in the Appendix gives more details of the token length distribution of all seven datasets. 
We can see that top-5 chunks can all fit into 4k sequence length (except few outliers) while top-10 and top-20 chunks can fit into 16k sequence length.

\subsection{Instruction Tuning}
To train the pretrained LLMs to follow instructions for question answering or text summarization, we also performed instruction tuning.
We first construct a blend of instruction tuning datasets consisting of 102K training samples from the Soda dataset~\citep{kim2022soda}, ELI5 dataset~\citep{fan2019eli5}, FLAN dataset~\citep{wei2021finetuned} , Open Assistatant dataset~\citep{kopf2023openassistant}, Dolly~\citep{DatabricksBlog2023DollyV2} and a proprietary sourced conversational dataset, to adapt all foundation models to follow instructions. 
In terms of the template, we use "System: \{System\}\textbackslash n\textbackslash nUser: \{Question\}\textbackslash n\textbackslash nAssistant: \{Answer\}" as the format to support multi-turn dialogue training. As all of the tasks contain the context information for reasoning over at inference time, we add the context before the dialogue, i.e. "System: \{System\}\textbackslash n\textbackslash n\{Context\}\textbackslash n\textbackslash nUser: \{Question\}\textbackslash n\textbackslash nAssistant: \{Answer\}".

We finetune the LLM by taking the loss only on the \{Answer\} part with batch size 128 and learning rate of 5e-6 for 1000 steps. 
 For the rest of the paper, results are all reported using the instruction tuned chat model on top of the foundational {\ourllm}, {Llama2-7B}, and {\llama}.

\begin{table}[t!]\small
\centering
\begin{tabular}{cccccccccc}
\toprule
 Model  & Seq len. & Avg.  & QM  & QASP & NQA & QLTY  &MSQ &HQA &MFQA \\
\midrule
{\ourllm} & 4k & 26.44 & 15.56 & 23.66 & 15.64 & 49.35 & 11.08 & 28.91 & 40.90\\ 
+ ret & 4k & 29.32 & 16.60 & 23.45 & 19.81 & 51.55 & 14.95 & 34.26 & 44.63\\ 
{\ourllm} & 16k & {29.45} & 16.09 & 25.75 & 16.94 & 50.05 & 14.74 & 37.48 & 45.08\\ 
+ ret & 16k & \textbf{29.65} & 15.69 & 23.82 & 21.11 & 47.90 & 15.52 & 36.14 & 47.39\\ 
\midrule
{\llama} & 4k & 31.61 & 16.34 & 27.70 & 19.07 & 63.55 & 15.40 & 34.64 & 44.55\\ 
+ ret  & 4k & {36.02}	& 17.41 & 28.74	& 23.41	& 70.15	& 21.39	& 42.06	& 48.96  \\
{\llama} & 16k & 36.78 & 16.72 & 30.92 & 22.32 & \textbf{76.10} & 18.78 & 43.97 & 48.63\\ 
+ ret & 16k & 37.23 & \textbf{18.70} & 29.54 & 23.12 & 70.90 & 23.28 & 44.81 & 50.24\\ 
{\llama} & 32k & 37.36 & 15.37 & \textbf{31.88} & 23.59 & 73.80 & 19.07 & 49.49 & 48.35\\ 
+ ret   & 32k & \textbf{39.60} & 18.34 & 31.27 & \textbf{24.53} & 69.55 & \textbf{26.72} & \textbf{53.89} & \textbf{52.91}\\ 
\midrule
Llama2-7B & 4k & 22.65  &  14.25  & 22.07 & 14.38  & 40.90 & 8.66 & 23.13  & 35.20 \\ 
+ ret  & 4k   & \textbf{26.04} &  16.45  & 22.97 & 18.18  & 43.25 & 14.68 & 26.62 & 40.10 \\
Llama2-7B  & 32k & \textbf{28.20} & 16.09 &  23.66  &  19.07  &  44.50 &  15.74 &  31.63 & 46.71 \\ 
+ ret   & 32k & 27.63 &  17.11  & 23.25  & 19.12  &  43.70 &  15.67 &  29.55 &  45.03\\ 

\bottomrule
\end{tabular}
\vspace{-.1cm}
\caption{Comparison of model variants ({\ourllm}, {Llama2-7B}, {\llama}) with sequence length ranging from 4k to 32k under seven datasets. ``ret'' denotes using the best retriever~(Dragon or Contriever or OpenAI embeddings) and here we used top-5 for the retriever. }
\vspace{-.1cm}
\label{tab:7datasets-comp}
\end{table}


\section{Results}
\label{sec:results}
In this section, we report the results and provide detailed analysis. 
\subsection{Main Results}
In Table \ref{tab:7datasets-comp}, we compare different model variants with context lengths ranging from 4K to as long as 32K using {\ourllm} and {\llama}.
First, we find that baseline models without retrieval of 4k sequence length achieve the worst results for both {\ourllm} and {\llama}. This is because the minimum average sequence length of all seven tasks exceeds 4096, the context window of the foundation models and therefore valuable texts get truncated randomly. 
As a result, retrieval is especially helpful for 4K LLMs e.g., {\llama}-4K is improved from 31.61 to 35.73 while {\ourllm}-4K is improved from 26.44 to 29.32. 
Second, we observe that HotpotQA (HQA) especially favors long sequence models as the score improves from 34.64 to 43.97 for {\llama} and from 28.91 to 37.48 for {\ourllm} when the sequence length increases from 4k to 16k. This is because Hotpot QA is a multi-hop dataset where the questions are not hard to answer but all intermediate hops are necessary to get correct answer. Therefore, long context are beneficial to increase the recall of incorporating all intermediate hops.

\begin{figure}[t]
  \centering
  \includegraphics[width=7cm]{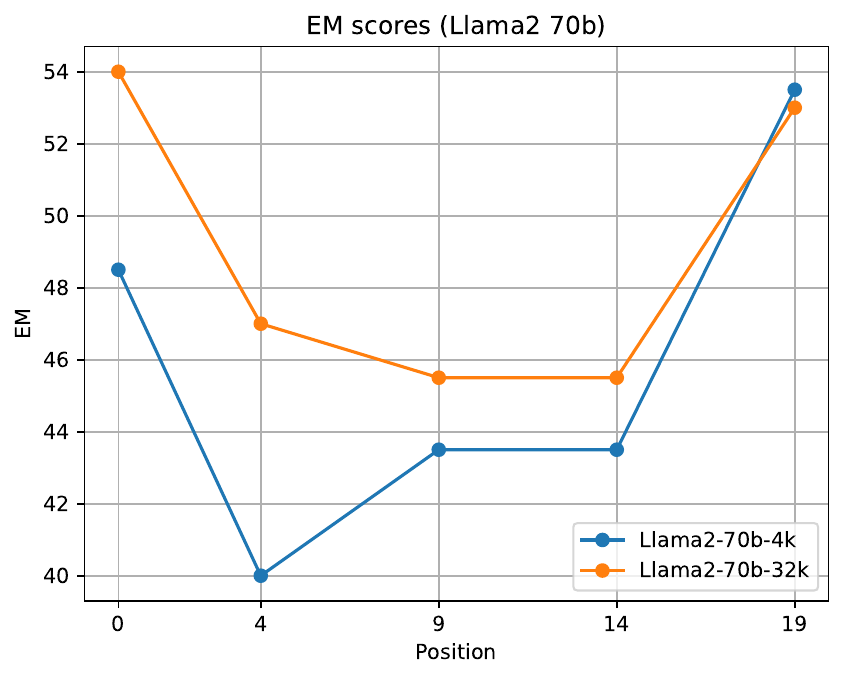}
    \vspace{-.3cm}
  \caption{{\llama} also displays lost-in-the-middle phenomenon}
  \label{fig:lost-in-the-middle}
\end{figure}

It is quite interesting that the retrieval-augmented long context LLM (e.g., 16K and 32K) can obtain better results than retrieval-augmented 4K context LLM, even they are feed with the same top 5 chunks of evidence. We hypothesize this interesting observation is related to the ``lost in the middle'' phenomenon~\citep{liu2023lost}, where the LLMs has such ``U-shaped'' performance curve. Specifically, LLMs are better at utilizing relevant information that occurs at the beginning or end of its input context window. To further verify the hypothesis, we conduct the “lost-in-the-middle” study following \citet{liu2023lost} for {\llama}-4k and {\llama}-32k. As show in Figure \ref{fig:lost-in-the-middle}, we confirm that the phenomenon also exists in {\llama} with different context lengths. In particular, the comparison of the curves from {\llama}-4k and {\llama}-32k suggests that the long context model has better accuracy for incorporating top-5 retrieved context.

\begin{table}[t!]
\centering
\resizebox{\textwidth}{!}{
\begin{tabular}{cclllllccccc} 
\toprule
Model   & Avg-7 & Avg-4* & QM*  & QASP* & NQA* & QLTY*  &MSQ &HQA &MFQA \\
\midrule
Davinci003~(175B)  & 39.2 & 40.8* & 16.9*	& 52.7*	& 24.6* & 69.0* & 22.1 & 41.2 &  47.8 \\
GPT-3.5-turbo~(4k)   & 38.4 & 39.2* & 15.6*  & 49.3* & 25.1* & 66.6* & 21.2 & 40.9 & 49.2 \\ 
+ret   &  &  &   &  &  &  &24.4  & 49.5 & 49.5 \\ 
GPT-3.5-turbo-16k &	42.8 &	42.4 &	17.6&	50.5	& 28.8	&72.6	&26.9&	51.6&	52.3 \\
+ret    &	 &	 &	&		& 	&	& 30.4 & 46.6 & 52.8	 \\
{\llama}-32k  & {40.9} & 42.4   &15.6		 & 45.9	 & 28.4	& 79.6   & 19.1 & 49.5 & 48.4 \\ 
{\llama}-32k-ret  & \textbf{43.6} & \textbf{43.0} & 18.5  & 46.3 & 31.5 & 75.6  & 26.7 & 53.9 & 52.9\\ 

\bottomrule
\end{tabular}
}
\caption{Comparison of our best retrieval-augmented {\llama}-32k-ret with GPT-3.5-turbo-16k and Davinci-003~(175B parameters).  For QMSum (QM), Qasper (QASP), NarrativeQA (NQA), QuALITY (QLTY), we used the test set from the ZeroSCROLLS leaderboard as the organizers have prepared the scores of GPT-3.5-turbo (4k) and Davinci-003~(highlighted with *). 
Avg-7 refers to the average score of all 7 datasets, and Avg-4* refers to the average of 4 datasets from ZeroSCROLLS.
}
\label{tab:compare-to-chatgpt}
\vspace{-2mm}
\end{table}

Note that, we have very different observation from the conclusion drawn from LongBench work~\citep{bai2023longbench}: \emph{``Retrieval brings improvement for model with weak ability on long contexts, but the performance still lags behind models that have strong long context understanding capability''}.
Here, we demonstrate retrieval can significantly improve the performance of both {\ourllm} and {\llama} regardless their context window size.
For example, our best retrieval-augmented {\llama}-32k-ret outperforms its baseline w/o retrieval by a margin, i.e., 39.60 vs. 37.36.
We think the major reason for such different conclusion is that \cite{bai2023longbench} uses much smaller LLM with 6B and 7B parameters, which usually has relatively worse zero-shot capability to incorporate the retrieved chunked context.
To further validate the hypothesis, we also report the results using  Llama2-7B in Table \ref{tab:llama2-70b-4k-16k-32k}. One can actually draw similar conclusions to \cite{bai2023longbench} . We think the underlying reasons are: i) For Llama2-7B-chat-4k, its short context length is the bottleneck for long context tasks. Thus, retrieval-augmentation largely improves the results. ii) For Llama2-7B-chat-32 and ChatGLM2-6B-32k, the context length bottleneck has been mostly removed. However, their retrieval-augmented models have limited zero-shot capability of incorporating retrieved chunks of context, due to the smaller size. As a result, retrieval is not helpful for both Llama2-7B-32k and ChatGLM2-6B-32k, which is different from large LLMs like Llama2-70B-32k in our case.

In contrast, the larger instruction tuned LLMs like {\llama} has much stronger zero-shot capability to incorporate retrieved evidence.
This observation is becoming more clear when one compares the gain of retrieval-augmentation between {\ourllm} and {\llama}, where {\llama} enjoys larger benefit of incorporating context through retrieval.

%

%


\subsection{Comparing to OpenAI models}
To further understand how good is our best model, i.e., augmenting {\llama}-32k with retrieval, we also compare it to GPT-3.5-turbo(4k),  GPT-3.5-turbo-16k and Davinci-003 on those seven datasets.\footnote{For QMSum (QM), Qasper (QASP), NarrativeQA (NQA), QuALITY (QLTY), we used the test set from the ZeroSCROLLS leaderboard as the organizers have prepared the scores of GPT-3.5-turbo~(4k) and Davinci-003 there.}  We found that {\llama}-32k-ret achieves better results than GPT-3.5-turbo-16k in terms of the average accuracy over seven datasets, while better than Davinci-003~(w/ 175B parameters) on the average over 4 tasks. This indicates {\llama}-32k with retrieval is a strong model for these long context tasks, and our conclusion is built on the state-of-the-art results. 

We also report the retrieval augmented results for GPT3.5-turbo on MSQ, HQA and MFQA. For GPT3.5-turbo-4k, retrieval significantly improves the performance (avg from 37.08 to 41.15). For GPT3.5-turbo-16k, the average scores for retrieval (43.27) and non-retrieval (43.60) scores are close to each other which are both lower than our Llam2-70B-32k-ret results (44.51). Note that GPT3.5-turbo-16k is a blackbox API, we don’t know how it is implemented, the model size as well as any preprocessing steps.

\begin{table}[t!]\small
\centering
\begin{tabular}{cccccccccc}
\toprule
Seq len & Setting &  Avg. & QM  & QASP & NQA & QLTY  &MSQ &HQA &MFQA \\
  \midrule
4k   & baseline (w/o ret) & 31.61 & 16.34 & 27.70 & 19.07 & 63.55 & 15.40 & 34.64 & 44.55\\   
 & Dragon & 35.73 & 18.14 & 29.20 & 23.39 & 70.30 & 20.09 & 41.54 & 47.45\\   
 & Contriever & \textbf{36.02}	& 17.41 & 28.74	& 23.41	& 70.15	& 21.39	& 42.06	& 48.96  \\
  & OpenAI-embedding  & 35.79	 & 17.76	 & 28.85	 & 23.57	 & 70.70	 & 19.92	 & 41.76	 & 47.99\\
 \midrule
32k & baseline (w/o ret) & 37.36 & 15.37 & {31.88} & 23.59 & 73.80 & 19.07 & 49.49 & 48.35\\   
 & Dragon & \textbf{39.60} & 18.34 & 31.27 & 24.53 & 69.55 & {26.72} & 53.89 & {52.91}\\   
 & Contriever  & 38.85	 & 17.60	 & 31.56	 & 23.88	 & 69.00	 & 26.61	 & 49.65	 & 53.66 \\
  & OpenAI-embedding   & 39.34	 & 18.24	 & 32.07	 & 24.36	 & 69.45	 & 24.90	 & 51.64	 & 54.75\\
\bottomrule
\end{tabular}
\caption{Comparisons of adding top 5 retrieved chunks from different retrievers to the context under {\llama}.}
\vspace{1mm}
\label{tab:compare-retriever}
\end{table}

\begin{table}[t!]\small
\centering
\begin{tabular}{cccccccccc}
\toprule
Seq len & Setting &  Avg. & QM  & QASP & NQA & QLTY  &MSQ &HQA &MFQA \\
  \midrule
4k  & base & 31.61 & 16.34 & 27.70 & 19.07 & 63.55 & 15.40 & 34.64 & 44.55\\   
 & top-5 & \textbf{35.73} & 18.14 & 29.20 & 23.39 & 70.30 & 20.09 & 41.54 & 47.45\\   
 & top-10 & 34.62 & 16.54 & 28.67 & 24.38 & 68.70 & 19.00 & 42.18 & 42.84\\   
 & top-20 & 34.61 & 16.52 & 28.67 & 24.38 & 68.70 & 19.00 & 42.18 & 42.84\\
 \midrule
16k & base & 36.78 & 16.72 & 30.92 & 22.32 & {76.10} & 18.78 & 43.97 & 48.63\\   
 & top-5 & 37.23 & {18.70} & 29.54 & 23.12 & 70.90 & 23.28 & 44.81 & 50.24\\   
 & top-10 & \textbf{38.31} & 18.41 & 30.20 & 25.53 & 73.60 & 22.78 & 47.72 & 49.91\\   
 & top-20 & 36.61 & 17.26 & 29.60 & 25.81 & 72.30 & 22.69 & 41.36 & 47.23\\   
 \midrule
32k & base & 37.36 & 15.37 & {31.88} & 23.59 & 73.80 & 19.07 & 49.49 & 48.35\\   
 & top-5 & \textbf{39.60} & 18.34 & 31.27 & 24.53 & 69.55 & {26.72} & 53.89 & {52.91}\\   
 & top-10 & 38.98 & 17.71 & 30.34 & {25.94} & 70.45 & 22.80 & {55.73} & 49.88\\   
 & top-20 & 38.38 & 16.36 & 30.42 & 24.42 & 69.60 & 24.51 & 54.67 & 48.65 \\
\bottomrule
\end{tabular}
\caption{Comparisons of adding top-5/10/20 retrieved chunks to the context under 4k, 16k, and 32k input sequence lengths using {\llama}. More context does not always give better results.}
\vspace{1.5mm}
\label{tab:llama2-70b-4k-16k-32k}
\end{table}

\subsection{Ablation on Different Retrievers}
To investigate the impacts of different retrievers on top of {\llama}, we compare Dragon, Contriever, and OpenAI embeddings on top of {\llama}-4k and  {\llama}-32k. 
The results in Table \ref{tab:compare-retriever} confirms that 
our finding, i.e., \emph{retrieval can boost the performance of both short context and long context LLMs}, is consistent across different retrievers. 

\subsection{Increasing the number of retrieved chunks}
To study the impact of adding more retrieved chunks to the context, we increase the number of retrieved chunks from 5 to 20 using Dragon retriever and the results can be found in Table \ref{tab:llama2-70b-4k-16k-32k}. We observe that for different sequence lengths, the best averaged results are obtained either from top 5 or top 10. Even if 20 chunks can still fit into the 16K and 32K context window~(as shown in Figure \ref{fig:7datasets-top5-ctxs-boxplot}), adding more chunks up to 20 is not helpful and will sometime hurt the performance. 
We believe this is related to the ``lost in the middle'' phenomenon~\citep{liu2023lost} or the model is getting distracted by irrelevant information and therefore needs further research. 

\subsection{Retrieval for Few-shot Tasks}

\begin{table}[t!]\small
\centering
\begin{tabular}{ccc}
\toprule
 Model     & Trec   & SAMSum \\
\midrule
GPT-3.5-turbo-16k  & 68 & 41.7\\
{\llama}  & 73   &  46.5 \\ 
{\llama}-ret  & 76 & 47.3 \\ 
\bottomrule
\end{tabular}
\vspace{-.1cm}
\caption{ Comparison of {\llama} to GPT-3.5-turbo-16k with two few-shot learning tasks from LongBench. Retrieval is helpful for few-shot learning as well. } 
\label{tab:few-shot-task}
\end{table}

In addition to the zero-shot tasks of query-based summarization tasks and question answering tasks mentioned above, we further investigate the benefits of long context models for few-shot tasks using two additional datasets (Trec and SAMSum) from LongBench. We take the question from each dataset as the query and use it to search relevant QA pairs provided in the given few-shot examples. Table \ref{tab:few-shot-task} shows that our best model {\llama}-32k-ret outperforms its non-retrieval {\llama}-32k baseline as well as GPT-3.5-turbo-16k by a large margin. It again confirms the benefits on using retrieval together with long context models.

\vspace{-.1cm}
\section{Conclusion}
\label{sec:conclusion}
\vspace{-.1cm}
In this work, we systematically study the retrieval-augmentation versus long context extension using the state-of-the-art LLMs after instruction tuning for various long context QA and query-based summarization tasks.
After study, we have the following interesting findings: 
\emph{i)} Retrieval largely boosts the performance of both 4K short context LLM and 16K/32K long context LLMs.
\emph{ii)} The 4K context LLMs with simple retrieval-augmentation can perform comparable to 16K long context LLMs, while being more efficient at inference.
\emph{iii)} After context window extension and retrieval-augmentation, the best model {\llama}-32k-ret can outperform GPT-3.5-turbo-16k and Davinci003 in terms of average score on a suit of downstream tasks with informative queries. 
Our study shed light on the promising direction of combining retrieval and long context techniques together to build better LLM.

\vspace{-.1cm}
\section{Future Directions}
\vspace{-.1cm}
There are many potential research directions that can be extended from this work. One direction is to develop advanced methods (e.g. memory or hierarchical attention) for existing pretrained large language models e.g. Llama2-70B, which is itself non-trivial. Also, further extending the context window to 64k and even longer would be a very interesting study for large 70B parameter models even though pre-training longer sequence requires much more computation. Lastly, how to mitigate the “lost-in-the-middle” phenomenon is an open research topic and continue pretraining with UL2 loss \citep{tay2022ul2} could be one potential solution.

\label{sec:future}

\bibliography{paper}

\begin{thebibliography}{75}
\providecommand{\natexlab}[1]{#1}
\providecommand{\url}[1]{\texttt{#1}}
\expandafter\ifx\csname urlstyle\endcsname\relax
  \providecommand{\doi}[1]{doi: #1}\else
  \providecommand{\doi}{doi: \begingroup \urlstyle{rm}\Url}\fi

\bibitem[Anthropic(2023)]{claud_100k}
Anthropic.
\newblock Introducing 100k context windows.
\newblock \url{https://www.anthropic.com/index/100k-context-windows}, 2023.

\bibitem[Bai et~al.(2023)Bai, Lv, Zhang, Lyu, Tang, Huang, Du, Liu, Zeng, Hou,
  et~al.]{bai2023longbench}
Yushi Bai, Xin Lv, Jiajie Zhang, Hongchang Lyu, Jiankai Tang, Zhidian Huang,
  Zhengxiao Du, Xiao Liu, Aohan Zeng, Lei Hou, et~al.
\newblock Longbench: A bilingual, multitask benchmark for long context
  understanding.
\newblock \emph{arXiv preprint arXiv:2308.14508}, 2023.

\bibitem[Beltagy et~al.(2020)Beltagy, Peters, and Cohan]{beltagy2020longformer}
Iz~Beltagy, Matthew~E Peters, and Arman Cohan.
\newblock Longformer: The long-document transformer.
\newblock \emph{arXiv preprint arXiv:2004.05150}, 2020.

\bibitem[Borgeaud et~al.(2022)Borgeaud, Mensch, Hoffmann, Cai, Rutherford,
  Millican, Van Den~Driessche, Lespiau, Damoc, Clark,
  et~al.]{borgeaud2022improving}
Sebastian Borgeaud, Arthur Mensch, Jordan Hoffmann, Trevor Cai, Eliza
  Rutherford, Katie Millican, George~Bm Van Den~Driessche, Jean-Baptiste
  Lespiau, Bogdan Damoc, Aidan Clark, et~al.
\newblock Improving language models by retrieving from trillions of tokens.
\newblock In \emph{ICML}, 2022.

\bibitem[Brown et~al.(2020)Brown, Mann, Ryder, Subbiah, Kaplan, Dhariwal,
  Neelakantan, Shyam, Sastry, Askell, et~al.]{brown2020language}
Tom Brown, Benjamin Mann, Nick Ryder, Melanie Subbiah, Jared~D Kaplan, Prafulla
  Dhariwal, Arvind Neelakantan, Pranav Shyam, Girish Sastry, Amanda Askell,
  et~al.
\newblock Language models are few-shot learners.
\newblock \emph{NeurIPS}, 2020.

\bibitem[Bulatov et~al.(2022)Bulatov, Kuratov, and
  Burtsev]{bulatov2022recurrent}
Aydar Bulatov, Yury Kuratov, and Mikhail Burtsev.
\newblock Recurrent memory transformer.
\newblock \emph{NeurIPS}, 2022.

\bibitem[Chen et~al.(2023)Chen, Wong, Chen, and Tian]{chen2023extending}
Shouyuan Chen, Sherman Wong, Liangjian Chen, and Yuandong Tian.
\newblock Extending context window of large language models via positional
  interpolation.
\newblock \emph{arXiv preprint arXiv:2306.15595}, 2023.

\bibitem[Child et~al.(2019)Child, Gray, Radford, and
  Sutskever]{child2019generating}
Rewon Child, Scott Gray, Alec Radford, and Ilya Sutskever.
\newblock Generating long sequences with sparse transformers.
\newblock \emph{arXiv preprint arXiv:1904.10509}, 2019.

\bibitem[Conover et~al.(2023)Conover, Hayes, Mathur, Xie, Wan, Shah, Ghodsi,
  Wendell, Zaharia, and Xin]{DatabricksBlog2023DollyV2}
Mike Conover, Matt Hayes, Ankit Mathur, Jianwei Xie, Jun Wan, Sam Shah, Ali
  Ghodsi, Patrick Wendell, Matei Zaharia, and Reynold Xin.
\newblock Free dolly: Introducing the world's first truly open
  instruction-tuned llm, 2023.
\newblock URL
  \url{https://www.databricks.com/blog/2023/04/12/dolly-first-open-commercially-viable-instruction-tuned-llm}.

\bibitem[Dai et~al.(2019)Dai, Yang, Yang, Carbonell, Le, and
  Salakhutdinov]{dai2019transformer}
Zihang Dai, Zhilin Yang, Yiming Yang, Jaime Carbonell, Quoc~V Le, and Ruslan
  Salakhutdinov.
\newblock Transformer-{XL}: Attentive language models beyond a fixed-length
  context.
\newblock In \emph{ACL}, 2019.

\bibitem[Dao(2023)]{dao2023flashattention}
Tri Dao.
\newblock Flashattention-2: Faster attention with better parallelism and work
  partitioning.
\newblock \emph{arXiv preprint arXiv:2307.08691}, 2023.

\bibitem[Dao et~al.(2022)Dao, Fu, Ermon, Rudra, and
  R{\'e}]{dao2022flashattention}
Tri Dao, Dan Fu, Stefano Ermon, Atri Rudra, and Christopher R{\'e}.
\newblock Flashattention: Fast and memory-efficient exact attention with
  io-awareness.
\newblock \emph{NeurIPS}, 2022.

\bibitem[Dasigi et~al.(2021)Dasigi, Lo, Beltagy, Cohan, Smith, and
  Gardner]{dasigi-etal-2021-dataset-qasper}
Pradeep Dasigi, Kyle Lo, Iz~Beltagy, Arman Cohan, Noah~A. Smith, and Matt
  Gardner.
\newblock A dataset of information-seeking questions and answers anchored in
  research papers.
\newblock In \emph{Proceedings of the 2021 Conference of the North American
  Chapter of the Association for Computational Linguistics: Human Language
  Technologies}, pp.\  4599--4610, Online, June 2021. Association for
  Computational Linguistics.
\newblock \doi{10.18653/v1/2021.naacl-main.365}.
\newblock URL \url{https://aclanthology.org/2021.naacl-main.365}.

\bibitem[Fan et~al.(2019)Fan, Jernite, Perez, Grangier, Weston, and
  Auli]{fan2019eli5}
Angela Fan, Yacine Jernite, Ethan Perez, David Grangier, Jason Weston, and
  Michael Auli.
\newblock Eli5: Long form question answering.
\newblock \emph{arXiv preprint arXiv:1907.09190}, 2019.

\bibitem[Gao et~al.(2021)Gao, Biderman, Black, Golding, Hoppe, Foster, Phang,
  He, Thite, Nabeshima, Presser, and
  Leahy]{DBLP:journals/corr/abs-2101-00027-pile-dataset}
Leo Gao, Stella Biderman, Sid Black, Laurence Golding, Travis Hoppe, Charles
  Foster, Jason Phang, Horace He, Anish Thite, Noa Nabeshima, Shawn Presser,
  and Connor Leahy.
\newblock The pile: An 800gb dataset of diverse text for language modeling.
\newblock \emph{CoRR}, abs/2101.00027, 2021.
\newblock URL \url{https://arxiv.org/abs/2101.00027}.

\bibitem[Guu et~al.(2020)Guu, Lee, Tung, Pasupat, and Chang]{guu2020retrieval}
Kelvin Guu, Kenton Lee, Zora Tung, Panupong Pasupat, and Mingwei Chang.
\newblock {REALM}: Retrieval augmented language model pre-training.
\newblock In \emph{ICML}, 2020.

\bibitem[He et~al.(2020)He, Fan, Wu, Xie, and Girshick]{He_2020_CVPR_MoCo}
Kaiming He, Haoqi Fan, Yuxin Wu, Saining Xie, and Ross Girshick.
\newblock Momentum contrast for unsupervised visual representation learning.
\newblock In \emph{Proceedings of the IEEE/CVF Conference on Computer Vision
  and Pattern Recognition (CVPR)}, June 2020.

\bibitem[Ho et~al.(2019)Ho, Kalchbrenner, Weissenborn, and
  Salimans]{ho2019axial}
Jonathan Ho, Nal Kalchbrenner, Dirk Weissenborn, and Tim Salimans.
\newblock Axial attention in multidimensional transformers.
\newblock \emph{arXiv preprint arXiv:1912.12180}, 2019.

\bibitem[Huang et~al.(2023)Huang, Ping, Xu, Shoeybi, Chang, and
  Catanzaro]{huang2023raven}
Jie Huang, Wei Ping, Peng Xu, Mohammad Shoeybi, Kevin Chen-Chuan Chang, and
  Bryan Catanzaro.
\newblock Raven: In-context learning with retrieval augmented encoder-decoder
  language models.
\newblock \emph{arXiv preprint arXiv:2308.07922}, 2023.

\bibitem[Izacard \& Grave(2021)Izacard and Grave]{izacard2021leveraging}
Gautier Izacard and {\'E}douard Grave.
\newblock Leveraging passage retrieval with generative models for open domain
  question answering.
\newblock In \emph{EACL}, 2021.

\bibitem[Izacard et~al.(2021)Izacard, Caron, Hosseini, Riedel, Bojanowski,
  Joulin, and Grave]{izacard2021contriever}
Gautier Izacard, Mathilde Caron, Lucas Hosseini, Sebastian Riedel, Piotr
  Bojanowski, Armand Joulin, and Edouard Grave.
\newblock Unsupervised dense information retrieval with contrastive learning,
  2021.
\newblock URL \url{https://arxiv.org/abs/2112.09118}.

\bibitem[Izacard et~al.(2022)Izacard, Lewis, Lomeli, Hosseini, Petroni, Schick,
  Dwivedi-Yu, Joulin, Riedel, and Grave]{izacard2022few}
Gautier Izacard, Patrick Lewis, Maria Lomeli, Lucas Hosseini, Fabio Petroni,
  Timo Schick, Jane Dwivedi-Yu, Armand Joulin, Sebastian Riedel, and Edouard
  Grave.
\newblock Few-shot learning with retrieval augmented language models.
\newblock \emph{arXiv preprint arXiv:2208.03299}, 2022.

\bibitem[Jiang et~al.(2022)Jiang, Gao, Araki, Ding, Wang, Callan, and
  Neubig]{jiang2022retrieval}
Zhengbao Jiang, Luyu Gao, Jun Araki, Haibo Ding, Zhiruo Wang, Jamie Callan, and
  Graham Neubig.
\newblock Retrieval as attention: End-to-end learning of retrieval and reading
  within a single transformer.
\newblock In \emph{EMNLP}, 2022.

\bibitem[Johnson et~al.(2019)Johnson, Douze, and J{\'e}gou]{faiss}
Jeff Johnson, Matthijs Douze, and Herv{\'e} J{\'e}gou.
\newblock Billion-scale similarity search with {GPUs}.
\newblock \emph{IEEE Transactions on Big Data}, 7\penalty0 (3):\penalty0
  535--547, 2019.

\bibitem[Kaiokendev(2023)]{kaiokendev2023}
Kaiokendev.
\newblock Things {I}'m learning while training {SuperHOT}.
\newblock \url{https://kaiokendev.github.io/til#extending-context-to-8k}, 2023.

\bibitem[Karpukhin et~al.(2020)Karpukhin, Oguz, Min, Lewis, Wu, Edunov, Chen,
  and Yih]{karpukhin2020dense}
Vladimir Karpukhin, Barlas Oguz, Sewon Min, Patrick Lewis, Ledell Wu, Sergey
  Edunov, Danqi Chen, and Wen-tau Yih.
\newblock Dense passage retrieval for open-domain question answering.
\newblock In \emph{EMNLP}, 2020.

\bibitem[Khandelwal et~al.(2019)Khandelwal, Levy, Jurafsky, Zettlemoyer, and
  Lewis]{khandelwal2019generalization}
Urvashi Khandelwal, Omer Levy, Dan Jurafsky, Luke Zettlemoyer, and Mike Lewis.
\newblock Generalization through memorization: Nearest neighbor language
  models.
\newblock \emph{arXiv preprint arXiv:1911.00172}, 2019.

\bibitem[Kim et~al.(2022)Kim, Hessel, Jiang, Lu, Yu, Zhou, Bras, Alikhani, Kim,
  Sap, et~al.]{kim2022soda}
Hyunwoo Kim, Jack Hessel, Liwei Jiang, Ximing Lu, Youngjae Yu, Pei Zhou,
  Ronan~Le Bras, Malihe Alikhani, Gunhee Kim, Maarten Sap, et~al.
\newblock {Soda}: Million-scale dialogue distillation with social commonsense
  contextualization.
\newblock \emph{arXiv preprint arXiv:2212.10465}, 2022.

\bibitem[Kitaev et~al.(2020)Kitaev, Kaiser, and Levskaya]{kitaev2020reformer}
Nikita Kitaev, {\L}ukasz Kaiser, and Anselm Levskaya.
\newblock Reformer: The efficient transformer.
\newblock In \emph{ICLR}, 2020.

\bibitem[K{\"o}pf et~al.(2023)K{\"o}pf, Kilcher, von R{\"u}tte, Anagnostidis,
  Tam, Stevens, Barhoum, Duc, Stanley, Nagyfi, et~al.]{kopf2023openassistant}
Andreas K{\"o}pf, Yannic Kilcher, Dimitri von R{\"u}tte, Sotiris Anagnostidis,
  Zhi-Rui Tam, Keith Stevens, Abdullah Barhoum, Nguyen~Minh Duc, Oliver
  Stanley, Rich{\'a}rd Nagyfi, et~al.
\newblock Openassistant conversations--democratizing large language model
  alignment.
\newblock \emph{arXiv preprint arXiv:2304.07327}, 2023.

\bibitem[Kočiský et~al.(2018)Kočiský, Schwarz, Blunsom, Dyer, Hermann,
  Melis, and Grefenstette]{10.1162/tacl_a_00023-narrativeqa}
Tomáš Kočiský, Jonathan Schwarz, Phil Blunsom, Chris Dyer, Karl~Moritz
  Hermann, Gábor Melis, and Edward Grefenstette.
\newblock {The NarrativeQA Reading Comprehension Challenge}.
\newblock \emph{Transactions of the Association for Computational Linguistics},
  6:\penalty0 317--328, 05 2018.
\newblock ISSN 2307-387X.
\newblock \doi{10.1162/tacl_a_00023}.
\newblock URL \url{https://doi.org/10.1162/tacl\_a\_00023}.

\bibitem[Lewis et~al.(2020)Lewis, Perez, Piktus, Petroni, Karpukhin, Goyal,
  K{\"u}ttler, Lewis, Yih, Rockt{\"a}schel, et~al.]{lewis2020retrieval}
Patrick Lewis, Ethan Perez, Aleksandra Piktus, Fabio Petroni, Vladimir
  Karpukhin, Naman Goyal, Heinrich K{\"u}ttler, Mike Lewis, Wen-tau Yih, Tim
  Rockt{\"a}schel, et~al.
\newblock Retrieval-augmented generation for knowledge-intensive nlp tasks.
\newblock \emph{NeurIPS}, 2020.

\bibitem[Lin(2004)]{lin-2004-rouge}
Chin-Yew Lin.
\newblock {ROUGE}: A package for automatic evaluation of summaries.
\newblock In \emph{Text Summarization Branches Out}, pp.\  74--81, Barcelona,
  Spain, July 2004. Association for Computational Linguistics.
\newblock URL \url{https://aclanthology.org/W04-1013}.

\bibitem[Lin et~al.(2023)Lin, Asai, Li, Oguz, Lin, Mehdad, Yih, and
  Chen]{lin2023train}
Sheng-Chieh Lin, Akari Asai, Minghan Li, Barlas Oguz, Jimmy Lin, Yashar Mehdad,
  Wen-tau Yih, and Xilun Chen.
\newblock How to train your dragon: Diverse augmentation towards generalizable
  dense retrieval.
\newblock \emph{arXiv preprint arXiv:2302.07452}, 2023.

\bibitem[Liu et~al.(2023)Liu, Lin, Hewitt, Paranjape, Bevilacqua, Petroni, and
  Liang]{liu2023lost}
Nelson~F Liu, Kevin Lin, John Hewitt, Ashwin Paranjape, Michele Bevilacqua,
  Fabio Petroni, and Percy Liang.
\newblock Lost in the middle: How language models use long contexts.
\newblock \emph{arXiv preprint arXiv:2307.03172}, 2023.

\bibitem[Liu et~al.(2018)Liu, Saleh, Pot, Goodrich, Sepassi, Kaiser, and
  Shazeer]{liu2018generating}
Peter~J Liu, Mohammad Saleh, Etienne Pot, Ben Goodrich, Ryan Sepassi, Lukasz
  Kaiser, and Noam Shazeer.
\newblock Generating wikipedia by summarizing long sequences.
\newblock In \emph{ICLR}, 2018.

\bibitem[Lo et~al.(2020)Lo, Wang, Neumann, Kinney, and
  Weld]{lo-etal-2020-s2orc}
Kyle Lo, Lucy~Lu Wang, Mark Neumann, Rodney Kinney, and Daniel Weld.
\newblock {S}2{ORC}: The semantic scholar open research corpus.
\newblock In \emph{Proceedings of the 58th Annual Meeting of the Association
  for Computational Linguistics}, pp.\  4969--4983, Online, July 2020.
  Association for Computational Linguistics.
\newblock \doi{10.18653/v1/2020.acl-main.447}.
\newblock URL \url{https://aclanthology.org/2020.acl-main.447}.

\bibitem[Mohtashami \& Jaggi(2023)Mohtashami and Jaggi]{mohtashami2023landmark}
Amirkeivan Mohtashami and Martin Jaggi.
\newblock Landmark attention: Random-access infinite context length for
  transformers.
\newblock \emph{arXiv preprint arXiv:2305.16300}, 2023.

\bibitem[Nakano et~al.(2021)Nakano, Hilton, Balaji, Wu, Ouyang, Kim, Hesse,
  Jain, Kosaraju, Saunders, et~al.]{nakano2021webgpt}
Reiichiro Nakano, Jacob Hilton, Suchir Balaji, Jeff Wu, Long Ouyang, Christina
  Kim, Christopher Hesse, Shantanu Jain, Vineet Kosaraju, William Saunders,
  et~al.
\newblock Web{GPT}: Browser-assisted question-answering with human feedback.
\newblock \emph{arXiv preprint arXiv:2112.09332}, 2021.

\bibitem[Nijkamp et~al.(2023)Nijkamp, Hayashi, Xie, Xia, Pang, Xia, and
  et~al.]{Nijkamp2023xgen}
Erik Nijkamp, Hiroaki Hayashi, Tian Xie, Congying Xia, Bo~Pang, Congying Xia,
  and et~al.
\newblock Long sequence modeling with {XGen}: A 7b {LLM} trained on 8k input
  sequence length.
\newblock \url{https://blog.salesforceairesearch.com/xgen/}, 2023.

\bibitem[OpenAI(2022)]{chatgpt}
OpenAI.
\newblock Introducing chatgpt.
\newblock \url{https://openai.com/blog/chatgpt}, 2022.

\bibitem[OpenAI(2023{\natexlab{a}})]{gpt4}
OpenAI.
\newblock Gpt-4.
\newblock \url{https://openai.com/research/gpt-4}, 2023{\natexlab{a}}.

\bibitem[OpenAI(2023{\natexlab{b}})]{openai_32k}
OpenAI.
\newblock Function calling and other {API} updates (longer context).
\newblock \url{https://openai.com/blog/function-calling-and-other-api-updates},
  2023{\natexlab{b}}.

\bibitem[Pang et~al.(2022)Pang, Parrish, Joshi, Nangia, Phang, Chen,
  Padmakumar, Ma, Thompson, He, and Bowman]{pang-etal-2022-quality}
Richard~Yuanzhe Pang, Alicia Parrish, Nitish Joshi, Nikita Nangia, Jason Phang,
  Angelica Chen, Vishakh Padmakumar, Johnny Ma, Jana Thompson, He~He, and
  Samuel Bowman.
\newblock {Q}u{ALITY}: Question answering with long input texts, yes!
\newblock In \emph{Proceedings of the 2022 Conference of the North American
  Chapter of the Association for Computational Linguistics: Human Language
  Technologies}, pp.\  5336--5358, Seattle, United States, July 2022.
  Association for Computational Linguistics.
\newblock \doi{10.18653/v1/2022.naacl-main.391}.
\newblock URL \url{https://aclanthology.org/2022.naacl-main.391}.

\bibitem[Parmar et~al.(2018)Parmar, Vaswani, Uszkoreit, Kaiser, Shazeer, Ku,
  and Tran]{parmar2018image}
Niki Parmar, Ashish Vaswani, Jakob Uszkoreit, Lukasz Kaiser, Noam Shazeer,
  Alexander Ku, and Dustin Tran.
\newblock Image transformer.
\newblock In \emph{ICML}, pp.\  4055--4064, 2018.

\bibitem[Press et~al.(2021)Press, Smith, and Lewis]{press2021train}
Ofir Press, Noah~A Smith, and Mike Lewis.
\newblock Train short, test long: Attention with linear biases enables input
  length extrapolation.
\newblock In \emph{ICLR}, 2021.

\bibitem[Radford et~al.(2019)Radford, Wu, Child, Luan, Amodei, Sutskever,
  et~al.]{radford2019language}
Alec Radford, Jeffrey Wu, Rewon Child, David Luan, Dario Amodei, Ilya
  Sutskever, et~al.
\newblock Language models are unsupervised multitask learners.
\newblock 2019.

\bibitem[Rae et~al.(2020)Rae, Potapenko, Jayakumar, and
  Lillicrap]{rae2019compressive}
Jack~W Rae, Anna Potapenko, Siddhant~M Jayakumar, and Timothy~P Lillicrap.
\newblock Compressive {T}ransformers for long-range sequence modelling.
\newblock In \emph{ICLR}, 2020.

\bibitem[Ratner et~al.(2023)Ratner, Levine, Belinkov, Ram, Magar, Abend,
  Karpas, Shashua, Leyton-Brown, and Shoham]{ratner2023parallel}
Nir Ratner, Yoav Levine, Yonatan Belinkov, Ori Ram, Inbal Magar, Omri Abend,
  Ehud Karpas, Amnon Shashua, Kevin Leyton-Brown, and Yoav Shoham.
\newblock Parallel context windows for large language models.
\newblock In \emph{ACL}, 2023.

\bibitem[Roy et~al.(2021)Roy, Saffar, Vaswani, and Grangier]{roy2021efficient}
Aurko Roy, Mohammad Saffar, Ashish Vaswani, and David Grangier.
\newblock Efficient content-based sparse attention with routing transformers.
\newblock \emph{Transactions of the Association for Computational Linguistics},
  2021.

\bibitem[Shaham et~al.(2022)Shaham, Segal, Ivgi, Efrat, Yoran, Haviv, Gupta,
  Xiong, Geva, Berant, and Levy]{shaham-etal-2022-scrolls}
Uri Shaham, Elad Segal, Maor Ivgi, Avia Efrat, Ori Yoran, Adi Haviv, Ankit
  Gupta, Wenhan Xiong, Mor Geva, Jonathan Berant, and Omer Levy.
\newblock {SCROLLS}: Standardized {C}ompa{R}ison over long language sequences.
\newblock In \emph{Proceedings of the 2022 Conference on Empirical Methods in
  Natural Language Processing}, pp.\  12007--12021, Abu Dhabi, United Arab
  Emirates, December 2022. Association for Computational Linguistics.
\newblock URL \url{https://aclanthology.org/2022.emnlp-main.823}.

\bibitem[Shaham et~al.(2023)Shaham, Ivgi, Efrat, Berant, and
  Levy]{shaham2023zeroscrolls}
Uri Shaham, Maor Ivgi, Avia Efrat, Jonathan Berant, and Omer Levy.
\newblock Zeroscrolls: A zero-shot benchmark for long text understanding.
\newblock \emph{arXiv preprint arXiv:2305.14196}, 2023.

\bibitem[Shi et~al.(2023)Shi, Min, Yasunaga, Seo, James, Lewis, Zettlemoyer,
  and Yih]{shi2023replug}
Weijia Shi, Sewon Min, Michihiro Yasunaga, Minjoon Seo, Rich James, Mike Lewis,
  Luke Zettlemoyer, and Wen-tau Yih.
\newblock Re{Plug}: Retrieval-augmented black-box language models.
\newblock \emph{arXiv preprint arXiv:2301.12652}, 2023.

\bibitem[Su et~al.(2021)Su, Lu, Pan, Murtadha, Wen, and Liu]{su2021roformer}
Jianlin Su, Yu~Lu, Shengfeng Pan, Ahmed Murtadha, Bo~Wen, and Yunfeng Liu.
\newblock Roformer: Enhanced transformer with rotary position embedding.
\newblock \emph{arXiv preprint arXiv:2104.09864}, 2021.

\bibitem[Tay et~al.(2020)Tay, Bahri, Yang, Metzler, and Juan]{tay2020sparse}
Yi~Tay, Dara Bahri, Liu Yang, Donald Metzler, and Da-Cheng Juan.
\newblock Sparse sinkhorn attention.
\newblock In \emph{ICML}, 2020.

\bibitem[Tay et~al.(2021)Tay, Bahri, Metzler, Juan, Zhao, and
  Zheng]{tay2021synthesizer}
Yi~Tay, Dara Bahri, Donald Metzler, Da-Cheng Juan, Zhe Zhao, and Che Zheng.
\newblock Synthesizer: Rethinking self-attention for transformer models.
\newblock In \emph{ICML}, 2021.

\bibitem[Tay et~al.(2022{\natexlab{a}})Tay, Dehghani, Bahri, and
  Metzler]{Tay2022efficient}
Yi~Tay, Mostafa Dehghani, Dara Bahri, and Donald Metzler.
\newblock Efficient transformers: A survey.
\newblock \emph{ACM Computing Surveys}, 2022{\natexlab{a}}.

\bibitem[Tay et~al.(2022{\natexlab{b}})Tay, Dehghani, Tran, Garcia, Wei, Wang,
  Chung, Bahri, Schuster, Zheng, et~al.]{tay2022ul2}
Yi~Tay, Mostafa Dehghani, Vinh~Q Tran, Xavier Garcia, Jason Wei, Xuezhi Wang,
  Hyung~Won Chung, Dara Bahri, Tal Schuster, Steven Zheng, et~al.
\newblock Ul2: Unifying language learning paradigms.
\newblock In \emph{The Eleventh International Conference on Learning
  Representations}, 2022{\natexlab{b}}.

\bibitem[Thakur et~al.(2021)Thakur, Reimers, R{\"u}ckl{\'e}, Srivastava, and
  Gurevych]{thakur2021beir}
Nandan Thakur, Nils Reimers, Andreas R{\"u}ckl{\'e}, Abhishek Srivastava, and
  Iryna Gurevych.
\newblock Beir: A heterogenous benchmark for zero-shot evaluation of
  information retrieval models.
\newblock In \emph{NeurIPS}, 2021.

\bibitem[Touvron et~al.(2023{\natexlab{a}})Touvron, Lavril, Izacard, Martinet,
  Lachaux, Lacroix, Rozi{\`e}re, Goyal, Hambro, Azhar,
  et~al.]{touvron2023llama}
Hugo Touvron, Thibaut Lavril, Gautier Izacard, Xavier Martinet, Marie-Anne
  Lachaux, Timoth{\'e}e Lacroix, Baptiste Rozi{\`e}re, Naman Goyal, Eric
  Hambro, Faisal Azhar, et~al.
\newblock Llama: Open and efficient foundation language models.
\newblock \emph{arXiv preprint arXiv:2302.13971}, 2023{\natexlab{a}}.

\bibitem[Touvron et~al.(2023{\natexlab{b}})Touvron, Martin, Stone, Albert,
  Almahairi, Babaei, Bashlykov, Batra, Bhargava, Bhosale,
  et~al.]{touvron2023llama2}
Hugo Touvron, Louis Martin, Kevin Stone, Peter Albert, Amjad Almahairi, Yasmine
  Babaei, Nikolay Bashlykov, Soumya Batra, Prajjwal Bhargava, Shruti Bhosale,
  et~al.
\newblock Llama 2: Open foundation and fine-tuned chat models.
\newblock \emph{arXiv preprint arXiv:2307.09288}, 2023{\natexlab{b}}.

\bibitem[Trivedi et~al.(2022)Trivedi, Balasubramanian, Khot, and
  Sabharwal]{trivedi-etal-2022-musique}
Harsh Trivedi, Niranjan Balasubramanian, Tushar Khot, and Ashish Sabharwal.
\newblock {M}u{S}i{Q}ue: Multihop questions via single-hop question
  composition.
\newblock \emph{Transactions of the Association for Computational Linguistics},
  10:\penalty0 539--554, 2022.
\newblock \doi{10.1162/tacl_a_00475}.
\newblock URL \url{https://aclanthology.org/2022.tacl-1.31}.

\bibitem[Tworkowski et~al.(2023)Tworkowski, Staniszewski, Pacek, Wu,
  Michalewski, and Mi{\l}o{\'s}]{tworkowski2023focused}
Szymon Tworkowski, Konrad Staniszewski, Miko{\l}aj Pacek, Yuhuai Wu, Henryk
  Michalewski, and Piotr Mi{\l}o{\'s}.
\newblock Focused transformer: Contrastive training for context scaling.
\newblock \emph{arXiv preprint arXiv:2307.03170}, 2023.

\bibitem[Wang et~al.(2023)Wang, Ping, Xu, McAfee, Liu, Shoeybi, Dong, Kuchaiev,
  Li, Xiao, et~al.]{wang2023shall}
Boxin Wang, Wei Ping, Peng Xu, Lawrence McAfee, Zihan Liu, Mohammad Shoeybi,
  Yi~Dong, Oleksii Kuchaiev, Bo~Li, Chaowei Xiao, et~al.
\newblock Shall we pretrain autoregressive language models with retrieval? a
  comprehensive study.
\newblock \emph{arXiv preprint arXiv:2304.06762}, 2023.

\bibitem[Wang et~al.(2022)Wang, Yang, Huang, Jiao, Yang, Jiang, Majumder, and
  Wei]{wang2022text}
Liang Wang, Nan Yang, Xiaolong Huang, Binxing Jiao, Linjun Yang, Daxin Jiang,
  Rangan Majumder, and Furu Wei.
\newblock Text embeddings by weakly-supervised contrastive pre-training.
\newblock \emph{arXiv preprint arXiv:2212.03533}, 2022.

\bibitem[Wang et~al.(2020)Wang, Li, Khabsa, Fang, and Ma]{wang2020linformer}
Sinong Wang, Belinda Li, Madian Khabsa, Han Fang, and Hao Ma.
\newblock Linformer: Self-attention with linear complexity.
\newblock \emph{arXiv preprint arXiv:2006.04768}, 2020.

\bibitem[Wei et~al.(2021)Wei, Bosma, Zhao, Guu, Yu, Lester, Du, Dai, and
  Le]{wei2021finetuned}
Jason Wei, Maarten Bosma, Vincent~Y Zhao, Kelvin Guu, Adams~Wei Yu, Brian
  Lester, Nan Du, Andrew~M Dai, and Quoc~V Le.
\newblock Finetuned language models are zero-shot learners.
\newblock \emph{arXiv preprint arXiv:2109.01652}, 2021.

\bibitem[Wei et~al.(2022)Wei, Tay, Bommasani, Raffel, Zoph, Borgeaud, Yogatama,
  Bosma, Zhou, Metzler, et~al.]{wei2022emergent}
Jason Wei, Yi~Tay, Rishi Bommasani, Colin Raffel, Barret Zoph, Sebastian
  Borgeaud, Dani Yogatama, Maarten Bosma, Denny Zhou, Donald Metzler, et~al.
\newblock Emergent abilities of large language models.
\newblock \emph{arXiv preprint arXiv:2206.07682}, 2022.

\bibitem[Xiong et~al.(2021)Xiong, Zeng, Chakraborty, Tan, Fung, Li, and
  Singh]{xiong2021nformer}
Yunyang Xiong, Zhanpeng Zeng, Rudrasis Chakraborty, Mingxing Tan, Glenn Fung,
  Yin Li, and Vikas Singh.
\newblock Nystr\"omformer: A nystr\"om-based algorithm for approximating
  self-attention.
\newblock \emph{AAAI}, 2021.

\bibitem[Yang et~al.(2018)Yang, Qi, Zhang, Bengio, Cohen, Salakhutdinov, and
  Manning]{yang-etal-2018-hotpotqa}
Zhilin Yang, Peng Qi, Saizheng Zhang, Yoshua Bengio, William Cohen, Ruslan
  Salakhutdinov, and Christopher~D. Manning.
\newblock {H}otpot{QA}: A dataset for diverse, explainable multi-hop question
  answering.
\newblock In \emph{Proceedings of the 2018 Conference on Empirical Methods in
  Natural Language Processing}, pp.\  2369--2380, Brussels, Belgium,
  October-November 2018. Association for Computational Linguistics.
\newblock \doi{10.18653/v1/D18-1259}.
\newblock URL \url{https://aclanthology.org/D18-1259}.

\bibitem[Yogatama et~al.(2021)Yogatama, de~Masson~d’Autume, and
  Kong]{yogatama2021adaptive}
Dani Yogatama, Cyprien de~Masson~d’Autume, and Lingpeng Kong.
\newblock Adaptive semiparametric language models.
\newblock \emph{Transactions of the Association for Computational Linguistics},
  9:\penalty0 362--373, 2021.

\bibitem[Zaheer et~al.(2020)Zaheer, Guruganesh, Dubey, Ainslie, Alberti,
  Ontanon, Pham, Ravula, Wang, Yang, et~al.]{zaheer2020big}
Manzil Zaheer, Guru Guruganesh, Avinava Dubey, Joshua Ainslie, Chris Alberti,
  Santiago Ontanon, Philip Pham, Anirudh Ravula, Qifan Wang, Li~Yang, et~al.
\newblock Big {B}ird: Transformers for longer sequences.
\newblock In \emph{NeurIPS}, 2020.

\bibitem[Zeng et~al.(2022)Zeng, Liu, Du, Wang, Lai, Ding, Yang, Xu, Zheng, Xia,
  et~al.]{zeng2022glm}
Aohan Zeng, Xiao Liu, Zhengxiao Du, Zihan Wang, Hanyu Lai, Ming Ding, Zhuoyi
  Yang, Yifan Xu, Wendi Zheng, Xiao Xia, et~al.
\newblock Glm-130b: An open bilingual pre-trained model.
\newblock \emph{arXiv preprint arXiv:2210.02414}, 2022.

\bibitem[Zhong et~al.(2021)Zhong, Yin, Yu, Zaidi, Mutuma, Jha, Awadallah,
  Celikyilmaz, Liu, Qiu, and Radev]{zhong-etal-2021-qmsum}
Ming Zhong, Da~Yin, Tao Yu, Ahmad Zaidi, Mutethia Mutuma, Rahul Jha,
  Ahmed~Hassan Awadallah, Asli Celikyilmaz, Yang Liu, Xipeng Qiu, and Dragomir
  Radev.
\newblock {QMS}um: A new benchmark for query-based multi-domain meeting
  summarization.
\newblock In \emph{Proceedings of the 2021 Conference of the North American
  Chapter of the Association for Computational Linguistics: Human Language
  Technologies}, pp.\  5905--5921, Online, June 2021. Association for
  Computational Linguistics.
\newblock \doi{10.18653/v1/2021.naacl-main.472}.
\newblock URL \url{https://aclanthology.org/2021.naacl-main.472}.

\bibitem[Zhu et~al.(2021)Zhu, Ping, Xiao, Shoeybi, Goldstein, Anandkumar, and
  Catanzaro]{zhu2021long}
Chen Zhu, Wei Ping, Chaowei Xiao, Mohammad Shoeybi, Tom Goldstein, Anima
  Anandkumar, and Bryan Catanzaro.
\newblock Long-short transformer: Efficient transformers for language and
  vision.
\newblock \emph{NeurIPS}, 2021.

\end{thebibliography}
\bibliographystyle{iclr2024_conference}

We show an example below where the smaller model Llama2-7B fails to incorporate relevant context, while larger models with retrieval could successfully predict the correct answer.

\appendix
\section{Appendix}
\begin{table*}[htpb!]
  \centering
  \begin{tabular}{l p{10cm}}
    \toprule
    \textbf{Chunk 1}     & On September 18, 2015, the deluxe edition of the album was released containing live and instrumental tracks from the standard edition album, in addition to the single "Light" featuring Little Dragon. Critical reception
... Angelspit has toured with Angel Theory, Ayria, Ikon, KMFDM, Tankt and The Crüxshadows, and have also shared the stage with bands such as The Sisters of Mercy, Nitzer Ebb, Skinny Puppy and Front Line Assembly. They performed with Lords of Acid during a 22-date U.S. tour in March 2011 and \textbf{toured the United States with Blood on the Dance Floor in October 2011}. History
Karl Learmont (ZooG) and Amelia Tan (Destroyx) met on an online zine forum. They shared an interest in zines and started the distro Vox Populis in 2002. \\ \midrule
\textbf{Chunk 2}     &  They then started making zines for themselves which became the lyrical inspiration for releases to follow. Angelspit was formed in 2003, and the duo then self-released their debut EP, Nurse Grenade on 3 October 2004. ... A video for the remix of "Sleep Now" was released on 2 October 2010. They released their third remix album, Carbon Beauty on 8 March 2011. This new remix album contains 3 new tracks as well as 10 remixes of tracks from the Hideous and Perfect album. A video for "Toxic Girl" was released on 13 April 2011, and a video for "Like It?
\\
    \midrule
\textbf{Chunk 3}     &   Passage 1:
Blood on the Dance Floor (band)
\textbf{Blood on the Dance Floor was an American electronic music group from Orlando, Florida, formed in 2006.} The group's longest standing lineup, from 2009 to 2016, consisted of Jesus David Torres also known as Dahvie Vanity (born 1984) and Jayy Von Monroe (born 1991). ... The CD was self-released in October 2008. Only 300 copies were made.Vanity and Ecstasy recorded the singles "Siq With a Q" and "Suicide Club" as a duo in 2008, and released three extended plays over the first half of 2009, I Scream I Scream, OMFG Sneak Peak, and Extended Play.
\\ \midrule
\textbf{Chunk 4}     &  title: , source: Lick It!" was released on 27 July 2011. On 15 April 2011, Angelspit announced the addition of three new members: guitarist Valerie Gentile (Black Tape for a Blue Girl, The Crüxshadows), drummer Chris Kling (Hanzel und Gretyl, Mortiis) and videojammer The Liar. The new line-up of Angelspit released their fourth studio album, Hello My Name Is on 11 October 2011. Matt James replaced Chris Kling in early 2012, and former Crüxshadows guitarist George Bikos filled in for Valerie Gentile on the band's 2012 tour...
\\ \midrule
\textbf{Chunk 5}     &  Vanity denied these allegations in a video.In April 2017, Vanity announced that there would be a new member, and that Blood on the Dance Floor would be returning on May 5 of that year without Jayy Von Monroe. ... On January 1, 2021, Vanity released a new version of Blood on the Dance Floor's "Bewitched" as "Bewitched Reimagined" featuring singer Sammy Beare. As of 2016, Jayy Von Monroe has continued to work as a drag monster under the name "The Dahli" and Vanity has continued to release music and merchandise by himself under both "The Most Vivid Nightmares" and "Dark Arts Official".
\\ \midrule
\textbf{Question:}     &  Angelspit toured the United States with an American electronic music duo from Orlando, Florida, that formed in what year? \\ \midrule
\textbf{LLaMA2-7b-32k}     & Angelspit toured the United States with Blood on the Dance Floor in October 2011. \\ 
\textbf{LLaMA2-7b-32k-ret}     & Angelspit toured the United States with Blood on the Dance Floor in October 2011. \\ 
\textbf{LLaMA2-70b-32k}     & 2011 \\ 
\textbf{LLaMA2-70b-32k-ret}     & 2006 \\ 
\textbf{Ground truth}     & 2006 \\ \bottomrule

  \end{tabular}
  \caption{All models fail to get the correct answer of 2006, except LLaMA2-70b-32k-ret, which shows how retrieval together with long context can help derive the correct answer. }
  \label{tab:comparison}
\end{table*}

\begin{figure}[t]
  \centering
  \includegraphics[width=13cm]{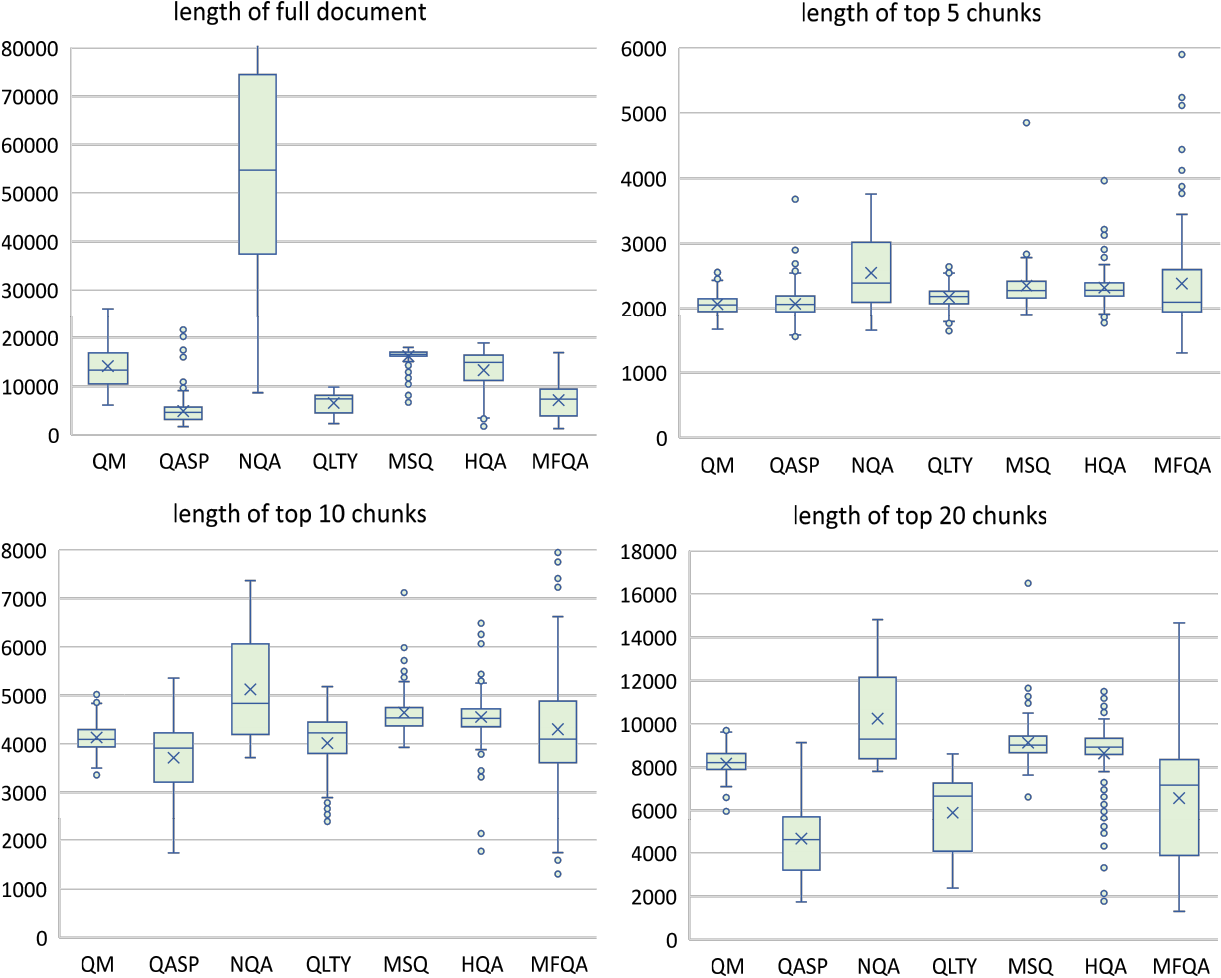}
  \caption{Token length distribution of the full document and the top-5, 10, 20 chunks of the seven datasets.}
  \label{fig:7datasets-top5-ctxs-boxplot}
\end{figure}

\end{document}